\documentclass[10pt,twocolumn,letterpaper]{article}

\usepackage[arxiv]{cvpr} 

\usepackage{xspace}
\usepackage{booktabs}
\usepackage{makecell}

\newcommand{\dname}{{SA-FARI}\xspace}
\newcommand{\samthree}{{SAM~3}\xspace}
\usepackage{changepage}

\newcommand{\mytablesize}{9}
\newcommand{\mytablebaselineskip}{11}

\newcolumntype{x}[1]{>{\centering\arraybackslash}p{#1pt}}
\newcolumntype{y}[1]{>{\raggedright\arraybackslash}p{#1pt}}

\renewcommand{\paragraph}[1]{
    \textbf{#1.}\noindent\xspace
}

\usepackage{pifont}
\newcommand{\cmark}{\ding{51}}
\newcommand{\xmark}{\ding{55}}
\usepackage{stfloats}

\newcommand{\cgf}{cgF\textsubscript{1}\xspace}
\newcommand{\pmf}{pmF\textsubscript{1}\xspace}
\newcommand{\ilmcc}{IL\_MCC\xspace}

\usepackage{etoolbox}
\pretocmd{\everypar}{\looseness=-1}{}{}

\newenvironment{absolutelynopagebreak}
  {\par\nobreak\vfil\penalty0\vfilneg
   \vtop\bgroup}
  {\par\xdef\tpd{\the\prevdepth}\egroup
   \prevdepth=\tpd}

\definecolor{cvprblue}{rgb}{0.21,0.49,0.74}
\usepackage[pagebackref,breaklinks,colorlinks,allcolors=cvprblue]{hyperref}
\usepackage{multirow}

\def\paperID{21259} \def\confName{CVPR}
\def\confYear{2026}

\title{The SA-FARI Dataset:\\Segment Anything in Footage of Animals for Recognition and Identification}

\author{
\begin{minipage}{\textwidth}
\centering
{\fontsize{9.5}{11.5}\selectfont
\vspace{-5pt}
Dante~Francisco~Wasmuht$^{1}$,
Otto~Brookes$^{3}$,
Maximillian~Schall$^{4}$,
Pablo~Palencia$^{5}$,
Chris~Beirne$^{6}$,
Tilo~Burghardt$^{3}$,
Majid~Mirmehdi$^{3}$,
Hjalmar~Kühl$^{7}$,
Mimi Arandjelovic$^{8}$,
Sam~Pottie$^{9}$,
Peter~Bermant$^{1}$,
Brandon~Asheim$^{1}$,
Yi~Jin~Toh$^{1}$,
Adam~Elzinga$^{1}$,
Jason~Holmberg$^{1}$,
Andrew~Whitworth$^{6}$,
Eleanor~Flatt$^{6}$,
Laura~Gustafson$^{2}$,
Chaitanya~Ryali$^{2}$,
Yuan-Ting~Hu$^{2}$,
Baishan~Guo$^{2}$,
Andrew~Westbury$^{2}$,
Kate~Saenko$^{2}$,
Didac~Suris$^{2}$\\[1.5ex]
$^{1}$Conservation~X~Labs~(CXL)\quad
$^{2}$Meta\quad
$^{3}$University~of~Bristol\quad
$^{4}$Hasso~Plattner~Institute\quad
$^{5}$University~of~Oviedo\quad
$^{6}$Osa~Conservation\quad
$^{7}$Senckenberg~Museum~of~Natural~History\quad
$^{8}$Max~Planck~Institute~for~Evolutionary~Anthropology\quad
$^{9}$Climate Corridors
}
\end{minipage}
}

\begin{document}

\crefname{section}{\S\!}{\S\S\!}
\Crefname{section}{\S\!}{\S\S\!}
\crefname{figure}{Fig.}{Figs.}
\Crefname{figure}{Fig.}{Figs.}
\crefname{table}{Tab.}{Tabs.}
\Crefname{table}{Tab.}{Tabs.}

\maketitle
\begin{abstract}
Automated video analysis is critical for wildlife conservation. A foundational task in this domain is multi-animal tracking (MAT), which underpins applications such as individual re-identification and behavior recognition. However, existing datasets are limited in scale, constrained to a few species, or lack sufficient temporal and geographical diversity -- leaving no suitable benchmark for training general-purpose MAT models applicable across wild animal populations. To address this, we introduce \dname, the largest open-source MAT dataset for wild animals. It comprises 11,609 camera trap videos collected over approximately 10 years (2014-2024) from 741 locations across 4~continents, spanning 99 species categories. Each video is exhaustively annotated 
culminating in $\sim$46 hours of densely annotated footage containing 16,224 masklet identities and 942,702 individual bounding boxes, segmentation masks, and species labels. Alongside the task-specific annotations, we publish anonymized camera trap locations for each video. Finally, we present comprehensive benchmarks on SA-FARI using state-of-the-art vision-language models for detection and tracking, including \samthree, evaluated with both species-specific and generic animal prompts. We also compare against vision-only methods developed specifically for wildlife analysis. SA-FARI is the first large-scale dataset to combine high species diversity, multi-region coverage, and high-quality spatio-temporal annotations, offering a new foundation for advancing generalizable multi-animal tracking in the wild. The dataset is available at \reviewhref{https://www.conservationxlabs.com/sa-fari}{conservationxlabs.com/SA-FARI}.
\end{abstract}

\vspace{-15pt}
\section{Introduction}
\label{sec:intro}

\begin{figure}[t]
            \centering
        \includegraphics[width=1.\linewidth]{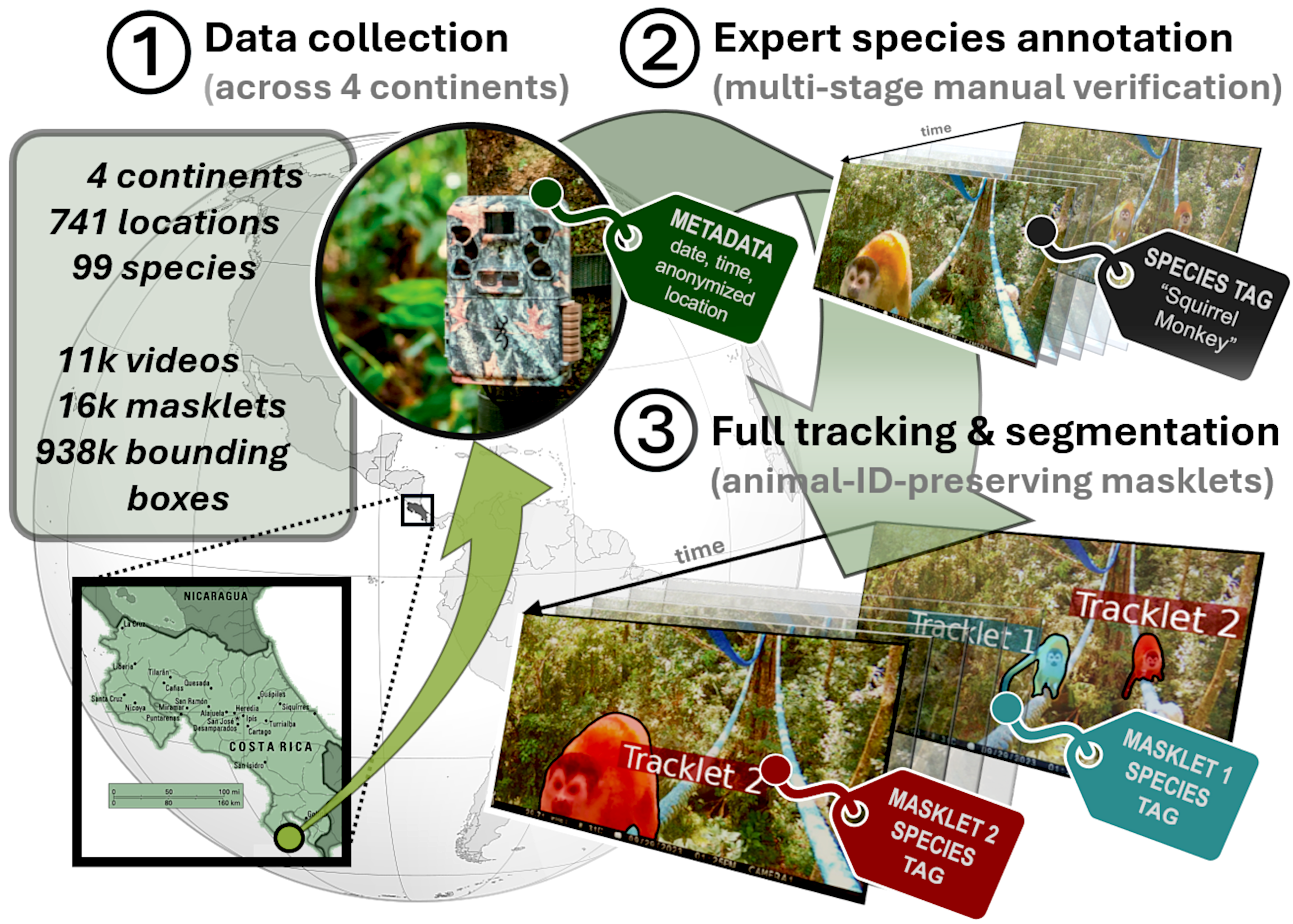}\vspace{-5pt}
                \caption{\textbf{SA-FARI Dataset Overview and Annotation.} We 1)~collect camera trap videos from 741 independent sampling locations across 4 continents, 2) label them with 99 species categories, and 3) exhaustively manually annotate spatio-temporal masklets for each individual animal.     Each video includes frame-level annotations,
    resulting in 16,224 unique identity masklets across $\sim$46 hours of video that form the by far largest dataset of its kind. Its rich annotations enable robust benchmarking of multi-animal tracking methods and support the development of generalizable, spatially accurate video understanding for wildlife.}
    \vspace{-12pt}
    \label{fig:dataset-overview}
\end{figure}
\vspace{-5pt}
\paragraph{Motivation} Biodiversity is declining at unprecedented rates -- orders of magnitude faster than at any time in the last tens of millions of years~\cite{ceballos2015accelerated}. To address this crisis, supportive tools are urgently needed to scale up conservation efforts~\cite{kuhl2013animal, tuia2022perspectives}. Among the most pressing is the development of automated video analysis methods~\cite{nathan2022big}, whose outputs (\eg species, individual identity, behaviour) are essential for downstream ecological and conservation-related tasks, such as abundance estimation, occupancy modelling, and health monitoring~\cite{chandler2013spatially, jimenez2019generalized, gilbert2021abundance}. The need for automated methods is particularly driven by the growing amount of data collected by in-situ sensors -- such as camera traps~(CTs) -- which now vastly exceed the processing capacity of human experts within the time frames required for effective conservation~\cite{burton2015wildlife, pollock2025harnessing}.

\begin{figure*}[]
    \centering
    \includegraphics[width=\textwidth]{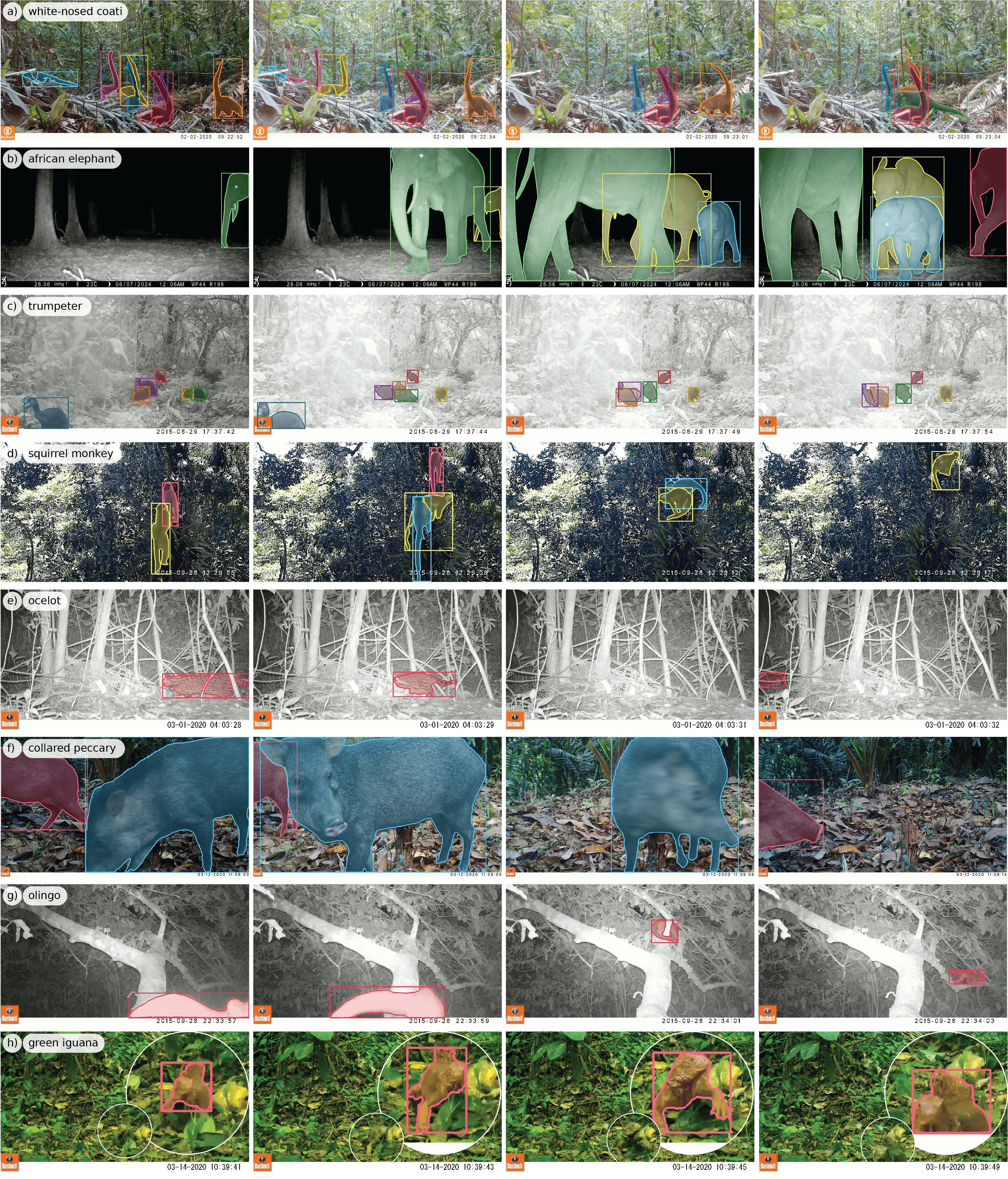}\vspace{-8pt}
                \caption{
        \textbf{Sample Skims from the \dname{} Dataset.} Each video–species pair is annotated with an exhaustive spatio-temporal segmentation of all animals belonging to that species category. The dataset captures a wide range of challenging scenarios, including: multiple animals in the same scene~(a–d), occlusions between animals or with other scene elements~(e), animals reappearing after leaving the frame~(f), unconventional or partial views~(g), small animals~(c), nighttime conditions~(b, e, g), and camouflaged animals~(h).
    }
    \label{fig:multiple}
\end{figure*}

\paragraph{Task} A foundational requirement for automated wild-life monitoring is the ability to localise individual animals in space and time, where localisation can for instance be achieved by either detection or segmentation in visual recordings. This task, jointly referred to as multi-animal tracking (MAT), underpins several critical computer vision applications, such as individual, species, and behaviour recognition~\cite{schofield2023automated, gadot2024crop, vogg2025computer}, by providing spatio-temporal animal representations that serve as a prerequisite for robust categorisation. Spatial representations encompass coat patterns, body shape, and pose~\cite{sharma2025universal,zuffi2024varen,shrack2025pairwise}, 
and also help to prevent shortcut learning of spurious background cues, which are known to hinder out-of-distribution~(OOD) generalisation~\cite{beery2018recognition, koh2021wilds, brookes2025panaf}. Temporal dynamics -- including gait, locomotion, and social interactions -- capture movement and behavioural context~\cite{baieri2025model, sakib2020visual, brookes2023triple}. Both spatial and temporal characteristics are unique to species, while also encoding information at the level of individuals and their behaviours.

\paragraph{State-of-the-art \& Limitations} Robust methodological progress is closely tied to the availability of datasets for system training and evaluation. In recent years, encouraging progress has been made, with several high-quality, large-scale animal datasets introduced for species classification~\cite{stevens2024bioclip}, and re-identification and behaviour recognition~\cite{ng2022animal, chen2023mammalnet, brookes2024panaf20k}. However, despite its recognised importance, MAT remains comparatively underdeveloped -- particularly for in-the-wild settings. While existing datasets are valuable, they exhibit several limitations: \begin{enumerate*}[label=(\roman*)]
\item Datasets developed specifically for MAT~\cite{zhang2023animaltrack, bai2021gmot, dave2020tao} lack scale 
~(less than one hour of annotated footage) and do not include videos captured by standard ecological sensors used in the field~(\eg CTs or unmanned areal vehicles (UAVs));
\item While several larger datasets do exist, they typically focus on a small number of species -- no more than 5 -- and even the most comprehensive among them provide no more than 10~hours of annotated footage~\cite{kholiavchenko2024kabr, Bain19}; \item In-the-wild tracking datasets based on UAV footage~\cite{kholiavchenko2024kabr, duporge2025baboonland, naik2024bucktales, dat2025wildlive} are generally collected from a single location or protected area. While they may span multiple habitats, they typically cover a narrow geographic range and short temporal window, limiting their ecological diversity; \item Finally, most datasets provide localisation only via bounding boxes. Where segmentation masks are included, they are often generated automatically without manual post-processing~\cite{dat2025wildlive}, leaving annotation quality uncertain \end{enumerate*}. These limitations mean that existing datasets are not suitable for training or evaluating urgently needed \textit{general} MAT models.

\paragraph{Contribution} We present \dname{}, by far the largest open-source MAT dataset for wild animals. \dname{} comprises 11,609 camera trap videos collected over 10 years from 741 locations across 4 continents. Each video is exhaustively annotated~(see \Cref{fig:dataset-overview}) with animal bounding boxes, segmentation masks, species category labels, and individual masklet identities, resulting in $\sim$46 hours of annotated footage spanning 99 wild animal species categories. This is 5$\times$ greater in total annotated duration and 2$\times$ broader in species diversity than any existing MAT dataset to date. Unlike prior datasets, \dname{} captures a wide range of species, behaviours, and environments across diverse ecological and geographical contexts, making it uniquely suited for training and evaluating \textit{general} MAT models. To our knowledge, it is also the only large-scale dataset to include high-quality manually annotated segmentation masks for wild multi-animal tracking. Additionally, we benchmark state-of-the-art video segmentation models on the \dname{} test set. 
Our results show that training on \dname{} leads to over 20-point gains in HOTA metrics, underscoring the value of our large-scale, diverse annotations.

\section{Related Work}
\label{sec:related_work}

\begin{table*}[!ht]\vspace{-11pt}
\centering
\small
\begin{tabular}{lrrrrrrrr}
\toprule
\textbf{Dataset} & \textbf{\# Sites} & \makecell{\textbf{\# Species}}  & \textbf{\# Videos} & \textbf{Duration (h)} & \textbf{\# Tracklets} & \textbf{Segmentation} & \textbf{Source} \\
\midrule
AnimalTrack \cite{zhang2023animaltrack}  & \textemdash{} & 10 & 58  & 0.23         & ~2k & \xmark & YouTube \\
GMOT \cite{bai2021gmot}$^\dagger$                  & \textemdash{} & 4  & 16  & 0.03         & 1k     & \xmark & YouTube \\
TAO \cite{dave2020tao} / BURST \cite{athar2023burst}$^\dagger$                 & \textemdash{} & 51 & 429  & 3.60          & 1k     & \xmark\ / \cmark & Varied\\
                  BuckTales \cite{naik2024bucktales}        & 1             & 1  & 12  & 0.21         & 0.7k     & \xmark & UAV\\
BaboonLaand \cite{duporge2025baboonland} & 1             & 1  & 18  & 0.32         & ~1k       & \xmark & UAV\\
KABR \cite{kholiavchenko2024kabr}        & 1             & 3  & 742 & 10.00        & \textemdash{}       & \xmark & UAV\\
WildLive \cite{dat2025wildlive}          & 1             & 3  & 22  &  0.70        & 0.3k     & \cmark & UAV\\
CCR \cite{Bain19}                              & 3             & 1  &  27   & 10.00  & 12k     & \xmark & Video Camera \\
MammAlps \cite{gabeff2025mammalps}       & 3             & 5  &   2,384  & 14.50       & ~6k      & \xmark & Camera Trap\\
PanAf500 \cite{brookes2024panaf20k}      & 12            & 2  & 500 & 2.10      & 1.5k & \xmark & Camera Trap \\
\textbf{\dname (ours)}                              & \textbf{741}             & \textbf{99}  &  \textbf{11,609}   & \textbf{45.78}  & \textbf{16k}     & \cmark & Camera Trap \\
\bottomrule
\end{tabular}
 \\ \ \\ \footnotesize{$\dagger$: For datasets that contain objects other than animals, we report numbers for the relevant fauna categories only.}\vspace{-5pt}
\caption{\textbf{A Comparison of Prominent Video Datasets for MAT}. \dname{} surpasses existing datasets in both diversity and scale, with nearly 100 species categories and hundreds of independent locations -- far exceeding other camera trap video datasets. It also provides more tracklets than any other dataset, including those beyond camera traps, and uniquely offers dense, manually verified segmentation annotations in form of accurate masklets. All datasets remain complementary, reflecting distinct ecological contexts and collection strategies.}\vspace{-8pt}
\label{tab:dataset_comparisons}
\end{table*}

MAT is a specific case of multi-object tracking (MOT), a well-established task in computer vision with numerous benchmark datasets and associated methods~\cite{dave2020tao, bai2021gmot}. While MOT has seen rapid progress in human-centric domains, development of MAT for wild animal monitoring has lagged behind -- largely due to the lack of large, diverse, and well-annotated datasets. In this section, we review existing resources (see Tab.~\ref{tab:dataset_comparisons}) and highlight their limitations regarding training generalisable models for wildlife conservation.

Throughout the paper, we use ``tracklet'' to refer to both \emph{bounding box} or \emph{mask} sequences that represent the temporal evolution of an object’s location or shape across frames. ``Masklets'' refer specifically to tracklets that contain mask information, \ie, sequences where each element is a segmentation mask rather than just a bounding box.

\paragraph{MAT Benchmark Datasets}
The AnimalTrack dataset~\cite{zhang2023animaltrack}, often cited as the primary MAT benchmark, contains 58 YouTube videos of animals in large groups (averaging 33 animals per video), with 1,927 annotated tracklets across 10 species. GMOT~\cite{bai2021gmot} includes 16 YouTube videos, with 980 tracklets over three species (averaging 61 animals per video). The TAO dataset~\cite{dave2020tao} comprises 429 videos from existing benchmarks (Charades~\cite{sigurdsson2016hollywood}, AVA~\cite{Gu2017AVAAV}, and HACS~\cite{zhao2019hacs}), with 985 annotated tracklets across 51 categories and two animals per video in average. BURST~\cite{athar2023burst} builds on top of TAO, annotating a segmentation mask for each bounding box. Other datasets include tracking annotations as a by-product of pose estimation or re-identification tasks. For instance, APT-36K~\cite{yang2022apt}, designed primarily for pose estimation, includes keypoint-based tracking annotations. Some datasets are collected in semi-controlled~\cite{waldmann20243d, ma2023chimpact} or fully controlled~\cite{naik20233d, pedersen20203d} environments such as enclosures or labs. While valuable, these datasets lack in-situ footage from ecological sensors, and therefore do not capture animals in natural environments, limiting generalisation to real-world conservation applications.

\paragraph{Wild MAT Datasets}
A small number of datasets focus on multi-animal tracking in the wild, primarily using UAV footage. BuckTales~\cite{naik2024bucktales} includes 12 high-resolution UAV videos of blackbuck antelopes, with 680 annotated tracklets and an average of 75 individuals per video. All footage was captured at a lekking site during the breeding season, limiting behavioural diversity. BaboonLand~\cite{duporge2025baboonland} contains 18 videos across three baboon troops, filmed in open landscapes at Mpala Research Centre, Kenya, with dense tracking annotations (up to 70 individuals per frame). KABR~\cite{kholiavchenko2024kabr, kholiavchenko2024deep} includes $\sim$10 hours of UAV footage of Grevy’s zebras, plains zebras, and giraffes, collected over 15 days at Mpala. WildLive~\cite{dat2025wildlive} features 22 drone videos from Ol Pejeta Conservancy, Kenya, with 291 tracklets of zebras, giraffes, and elephants, and automatically generated segmentation masks. While these datasets contain valuable in-the-wild footage, they are limited to a narrow geographic and temporal scope (typically a single protected area over short deployments), focus on only a few species, and often capture animals within a single behavioural context such as mating or social aggregation. Additionally, they rely heavily on automated annotations without manual correction, limiting the quality of spatio-temporal annotations.

\paragraph{Behaviour \& Re-ID Datasets with MAT Annotations}
Several high-quality datasets not originally developed for tracking include MAT annotations to support behaviour or identity recognition. LoTE~\cite{liu2023lote} contains 28k images with bounding boxes and segmentation masks for behaviour recognition across 11 species. MammAlps~\cite{gabeff2025mammalps} includes 6K tracklets spanning 8.5 hours of footage from nine cameras across three alpine sites, covering five species (over 90\% of footage features deer). PanAf500~\cite{brookes2024panaf20k} provides bounding boxes and tracklets for 500 camera trap videos of chimpanzees and gorillas (180k frames, $\sim$2 hours). The Chimpanzee Faces and Re-identification (CCR) dataset~\cite{Bain19} includes over 5k face and 12k body tracks (1M frames, $\sim$10 hours) from three field sites. The Rolandseck and Bavarian Forest datasets~\cite{Schindler_et_al_2024} contain dense tracking and segmentation annotations for less than 100 videos but are not publicly available. These datasets, while valuable, lack the species coverage, spatio-temporal breadth, and annotation precision necessary to support the development of generalisable MAT models for wildlife conservation.

\section{The \dname Dataset}\vspace{-5pt}
\label{sec:dataset}

\paragraph{Data Collection}
To create \dname we consolidated a large number of camera trap videos into one large public dataset. Data came from 4 continents, representing distinct ecoregions (Central Africa, South America, Mesoamerica, Southern Europe; see \Cref{fig:datasets}) and seven partner organizations: \anonymize{Osa Conservation, Los Amigos Biological Station, Institute for Game and Wildife Research (IREC, UCLM-CSIC-JCCM), Biodiversity Research Institute (IMIB, UO-CSIC-PA), Pan African Programme: The Cultures Chimpanzee and Conservation X Labs}. 
The camera trap videos were recorded across 741 independent sampling points. 

Data was collected using several brands and models of commercially available camera traps (Browning, Bushnell, Reconyx, Solaris, Spypoint, Ltl Acorn, Tetrao). Camera traps were set up using standardized procedures usually 0.3m to 2m from the ground at known wildlife crossings, gathering, feeding, resting or potential viewing sites. 2,790 videos from Costa Rica were recorded by cameras set up in tree crowns (arboreal camera traps) between 8 and 24m from the ground. Camera traps are motion triggered, and at night time videos are recorded with an IR flash invisible to the species of interest. Videos were captured at different resolutions (320$\times$194 to 2688$\times$1234) with frame rates ranging from 10 to 60 FPS (while most were captured at 30~FPS) and videos lasting between 0.5s and 90s (with most videos lasting 15s). The total data collection spanned approximately 10 years (2014-2024). All videos within the dataset contain at least one animal. All camera trap videos also contain audio (not used in this work).

\begin{figure}[!ht]\vspace{-5pt}
    \centering
    \includegraphics[width=\linewidth]{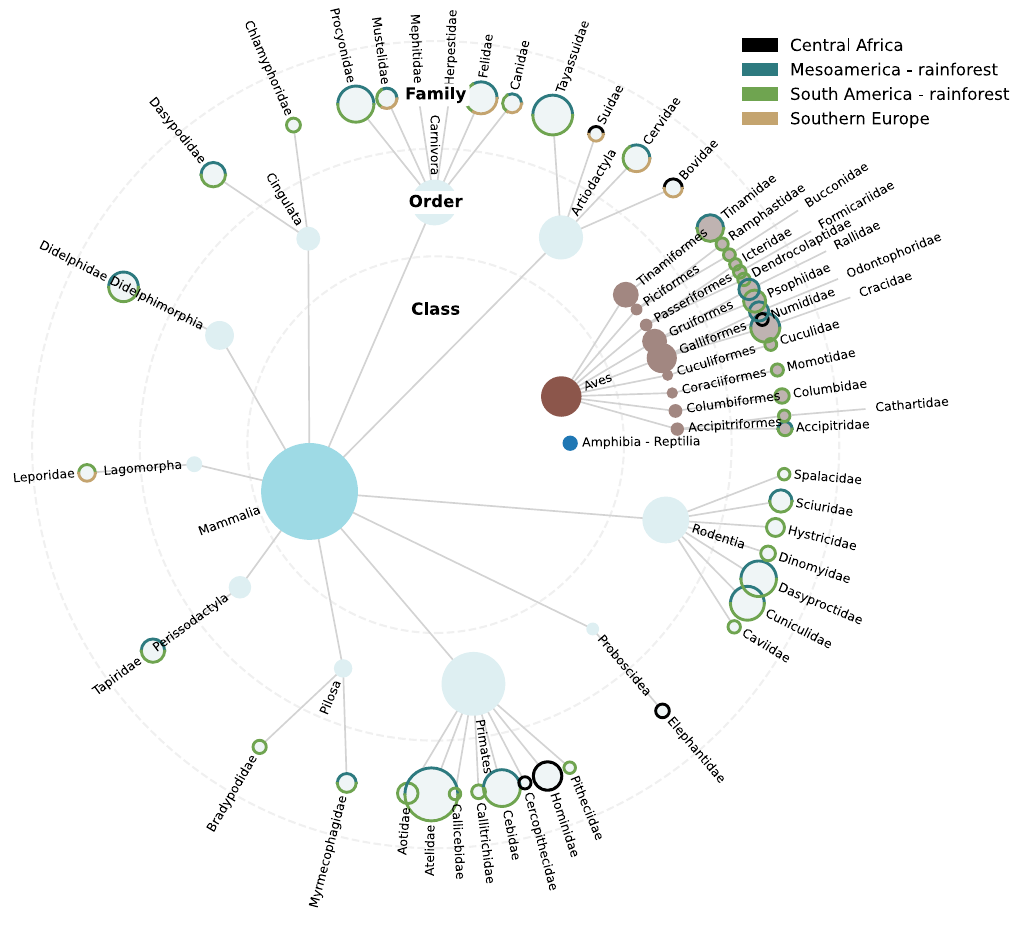}\vspace{-5pt}
    \caption{\textbf{Taxonomic Abundance of Videos.} Each circle represents the relative abundance (\ie number of videos) of taxonomic groups within the \dname{} dataset, from Class to Order to Family. Circle size is proportional to the number of associated videos, while edge colors indicate the continent(s) where the taxa were recorded. This visualisation highlights both the taxonomic and geographic diversity captured in the dataset.}\vspace{-15pt}
    \label{fig:order_taxa}
\end{figure}

\begin{figure}[!hb]
    \centering
    \includegraphics[width=1.\linewidth]{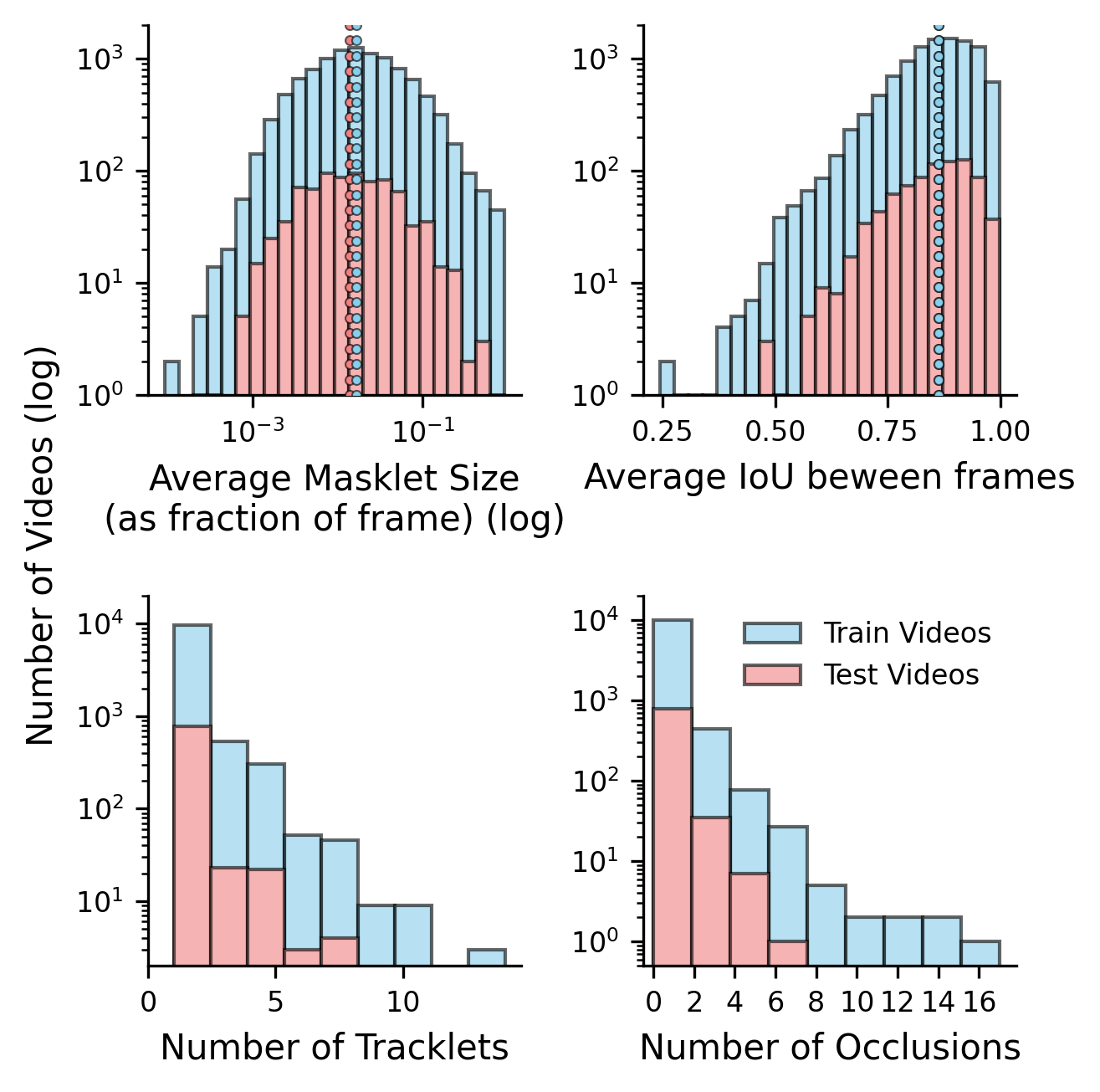}
        \caption{\textbf{Masklet Statistics.} Distribution of key masklet-level metrics across the \dname{} dataset. Average masklet size and average IoU are computed by first averaging within each masklet and then across all masklets in a video. The total number of masklets and occlusion events are summed per video. Dotted lines indicate the medians of each distribution.}
    \label{fig:tracklets}
\end{figure}

\paragraph{Species Annotation}
Each video was annotated with common names (\eg fox) for animals observed in the video by several experts from the local teams that provided the data. For each video, 2-4 independent annotators then confirmed the common name of the observed animals. Videos with animals of different species were subsequently disambiguated during the temporal segmentation and tracklet annotation process by assigning independent common names to each tracklet. Common names of animals are referred to as species categories elsewhere in this paper. 

We also provide the latin names (\eg \textit{Vulpes vulpes}) and associated taxonomic ranks and hierarchy (\eg Kingdom: Animalia - Phylum: Chordata - Class: Mammalia - Order: Carnivora - Family: Canidae - Genus: \textit{Vulpes} - Species: \textit{Vulpes vulpes}) down to at least the family but usually down to the species level, where appropriate. See Fig.~\ref{fig:order_taxa} for an overview of the species distribution statistics. For each video we also provide date and time stamps (if available) as well as an anonymized string indicating the sampling site. 

\begin{figure*}[!h]\vspace{-11pt}
    \centering
    \includegraphics[width=\linewidth]{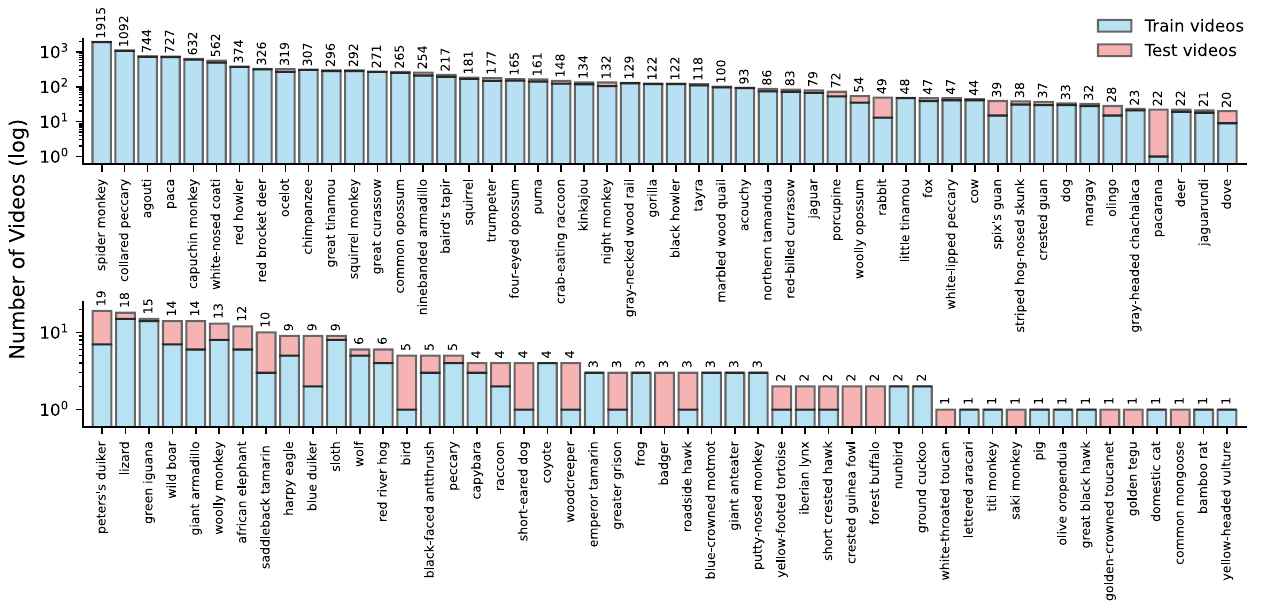}\vspace{-8pt}
    \caption{
        \textbf{Distribution of Species Category in the \dname{} Dataset.} The two panels show the number of videos per species category, broken down by data split. The distribution follows a long-tailed pattern typical of real-world wildlife datasets, with a few dominant species and many rarely observed ones. Notably, several species, such as the Saki monkey, appear only in the test set, reflecting the natural open-world setting of camera trap deployments.
    }
    \label{fig:species_stats}\vspace{-11pt}
\end{figure*}

\paragraph{Segmentation Mask Annotation}
The spatio-temporal segmentation annotations were created as part of the \samthree\reviewfootnote{\label{fn:samthree}\samthree is part of a parallel submission and therefore cannot be formally cited.} data engine, which annotates the videos at 6 fps. 
The process begins with an automatic stage, where \samthree is used to pseudo-annotate all video-species category pairs, generating initial masks for each sample. These masklets are then deduplicated based on their Intersection over Union (IoU) to remove redundant annotations.

Following the automatic step, human annotators review the results. 
They first check that the animal is visible and can be segmented, excluding cases where it is too blurred or indistinct (\eg within a flock). Next, a second annotator corrects the masklets by removing incorrect ones and using online SAM~2 \citep{sam2} in the loop to improve or add masklets as needed. Finally, a round of exhaustivity checking is performed to confirm that all possible masklets have been included. Bounding boxes are not annotated manually, they are trivially derived from the segmentation masks by computing their spatial limits. For further details on the human annotation process, see the \samthree~\citep{sam3} paper.

\paragraph{Train \& Test Splits}
The dataset was divided into training and test subsets according to the following procedure.
We set a maximum of 1,000 videos for the test set. Within this limit, we aimed to maximize the diversity of species categories represented, while ensuring that all videos from the same camera trap are assigned to the same split.
To achieve this, camera traps were sorted by their ratio of species categories to videos. We then greedily selected camera traps in order, prioritizing those that contributed the most new species categories to the test set. After each selection, the ordering was updated to deprioritize camera traps whose species categories had already been included in the test set. This process continued until the test set reached the target size.
This approach ensures a diverse and representative test set, while maintaining strict separation of camera trap locations between the training and test splits. Fig.~\ref{fig:species_stats} shows the full species distribution statistics across splits.

\paragraph{Category Augmentation} Next, we augment the dataset with \textit{negative} species category labels, which are essential for evaluating and training the precision of detection models. Negative species are those that do \emph{not} appear in a given video. We generate two types of negative species labels, each with a different level of difficulty. First, we group species categories into families, and if a family contains very few species categories, we further group them into orders or classes, resulting in a total of 29 groups. For every video-species category pair, we randomly select another species category from the same group as a hard negative (ensuring that it is not present in the video), as well as a species category from a different group as an easy negative (also randomly selected and not present in the video). This approach is feasible because the label annotation is exhaustive, meaning that all species categories present in each video are fully annotated. 

\paragraph{Test Set Partitioning} We further partition the \dname test set into five subsets: \emph{challenging}, \emph{night}, \emph{multi-masklet}, \emph{large-masklet}, and \emph{small-masklet}. These subsets are intended to enable a more detailed analysis of the dataset characteristics and the performance of different methods. 
A masklet is considered ``challenging'' if it meets at least one of the following criteria: (1) the animal is moving, as measured by the average inter-frame IoU of the masklet being $<0.7$ (\ie, when the overlap between the masklet's masks in consecutive frames is low, indicating significant movement in pixel coordinates), or (2) the number of occlusions is $\geq 2$. The challenging subset (106 videos) includes all videos that contain at least one such masklet. The night subset (598 videos) includes all the videos recorded at night. The multi-masklet subset (136 videos) consists of videos with two or more animals, regardless of species category. Finally, we define the large-masklet subset (211 videos) as the set of videos whose average masklet size is greater than the 75th percentile, and the small-masklet subset (212 videos) as those with average masklet size smaller than the 25th percentile. The masklet size is defined as the average mask size per frame, calculated over all frames in which the masklet is present. Each frame-level mask size is normalized by the image size, resulting in a value between 0 and 1. \Cref{fig:multiple} shows examples from the \dname{} dataset, including videos with multiple masklets, examples of occlusions, and other challenging scenarios.

\begin{table*}[t]
\centering
\small
\begin{tabular}{lrrrrrrr}
\toprule
 & \makecell{\textbf{\# Videos} \\ {} }
 & \makecell{\textbf{Duration}\\{(min)}}
 & \makecell{\textbf{\# Species} \\ {}} 
 &  \makecell{\textbf{\# Masklets} \\ {}}
 & \makecell{\textbf{\# Annotations}\\{(boxes \& masks)}} 
 & \makecell{\textbf{\# Video-species pairs} \\ {(incl. negatives)}} 
 & \makecell{\textbf{\# Sampling Locations} \\ (independent sites)} \\
\midrule
\textbf{Train} & 10,776 & 2,545 & 91 & 15,141 & 880,361 & 31,282 & 650 \\
\textbf{Test} & 833 & 202 & 83 & 1,083 & 62,341 & 2,322 & 91 \\
\midrule
\textbf{Total} & 11,609 & 2,747 & 99 & 16,224 & 942,702 & 33,604 & 741 \\
\bottomrule
\end{tabular}
\caption{\textbf{Dataset and Split Statistics.} 
Summary of key statistics for the \dname{} dataset across training and test splits. The test split is designed to maximise both species diversity and site diversity, with no overlap in camera trap locations between splits. Metrics include total number of videos, annotated duration (in minutes), species categories, spatio-temporal masklets, annotated bounding boxes and segmentation masks, video–species pairs (including negatives), and the number of independent sampling sites.}\vspace{-5pt}
\label{tab:dataset_statistics}
\end{table*}

\paragraph{Dataset Statistics}
\dname contains 11,609 videos of 99 different animal species categories (4 classes, 23 orders, 53 families and 82 genera). 29 species categories contribute to 90\% of the data with spider monkeys, collared peccaries and agoutis being the top-3 most common species (see \Cref{fig:species_stats}). 67 (67.7\%) of the species categories belong to the Mammalia class with 10 orders, 33 families and 57 genera; 27 (27\%) species categories belong to the Aves class with 10 orders, 17 families and 23 genera; 4 (4.0\%) species categories belong to the Reptilia class with 2 orders, 2 Families and 2 genera. 1 (1.0\%) species category belongs to the Amphibia class (see \Cref{fig:species_stats}). The most data and largest variety in species categories came from the South- and Mesoamerican recording locations. A minimum of 165 (22.3\%) recording locations provided 90\% of the data (see \Cref{fig:ranked}) and a minimum of 147 (19.8\%) recording locations provided 90\% of all species category variety. 

The dataset contains 16,224 masklets, \ie individual animals, while each video contains on average of 1.4 masklets, with 2,623 videos (22.7\%) containing 2 masklets or more and some videos containing up to 14 masklets. The top three species categories with the highest average number of masklets/video were red river hogs (6.7 masklets/video), white-lipped peccaries (3.5 masklets/video) and african elephants (3.3 masklets/video). The distributions of average masklet sizes, object motion measured via interframe IoU, number of masklets and masklet occlusions on the video level are shown in \Cref{fig:tracklets}.

\section{Benchmarks}
\label{sec:experiments}

In this section, we benchmark a range of animal detection and retrieval methods on the \dname{} dataset. In particular, we evaluate two classes of models: (1) vision–language models for the spatio-temporal 
localisation of species category instances in video, based on species name prompting, and (2) species-agnostic animal instance detection using vision-only generic detectors combined with strong general-purpose tracking algorithms. Note that the latter can be simulated via prompting of vision–language models using a generic query (\eg ``animal'') to align with more common vision-only benchmarks in animal tracking.

\paragraph{Species-Specific Prompt Evaluation}
In this setting, all vision–language models are provided with the species category via a prompt and are tasked to localise (\ie detect or segment) and track all instances of that species throughout the video. Specifically, we follow the evaluation protocol for video promptable concept segmentation (PCS) described in \samthree~\cite{sam3}.

Overall, we benchmark three models. The first two are GLEE~\cite{glee}, an open-vocabulary detector and tracker, and LLMDet~\cite{fu2025llmdet}, a recent detection-only model designed for open-world scenarios. Since LLMDet lacks an integrated tracker, we pair it with an independent tracking module. 

The third model, \samthree, extends the Segment Anything Model 2 (SAM 2)~\cite{sam2} to support open-vocabulary, text-based promptable segmentation and tracking. It can segment and track all instances of a user-specified concept across images and videos, using a DETR-style detector and a SAM 2-style memory-based tracker with a shared vision-language backbone. \samthree is trained on a large, diverse dataset of images and videos annotated with millions of unique concept phrases, collected via a human- and AI-in-the-loop pipeline. It is currently the state-of-the-art model for open-vocabulary segmentation and tracking across a wide range of domains.

As shown in \Cref{tab:results}, \samthree outperforms both GLEE and LLMDet, exceeding the performance of the latter~(row 2), closest competing model by +11.4 \cgf and + 7.2 pHOTA. Noting \samthree's superior performance in this setting, we evaluate \samthree under two additional configurations: (1) training with \dname{} included proportionally alongside other datasets, and (2) explicitly fine-tuned on \dname{}. The three variants are referred to as \samthree, \samthree~(\dname), and \samthree~FT~(\dname), respectively.

\samthree achieves the highest performance when it is fine-tuned on \dname~(row 5). Notably, incorporating \dname data into training (row 4) or fine-tuning~(row 5) leads to substantial improvements in performance. Specifically, the fine-tuned \samthree model~(row 5) outperforms its counterpart~(row 3) by +32.9, +19.6, and +19.1 on \cgf pHOTA, and TETA, respectively, exhibiting the value of domain-specific data on the performance of SOTA models.

\paragraph{Species-Agnostic Prompt Evaluation}
In this setting, species categories are replaced with the generic query ``animal''. While this is not the primary intended use case for the benchmark, it reflects common practice in the camera trap detection and tracking literature~\cite{beery2019efficient, gabeff2025mammalps, Schindler_et_al_2024} and, thus, can provide comparative context in a well-established domain. For this evaluation we utilise our test set videos with exhaustive species annotation with standard F1 and HOTA metrics. Results are reported for the \samthree (\dname) configuration explained above, as well as for three tracking methods: BoostSort++ \cite{stanojevic2024btpp}, OCSort \cite{cao2023observation}, and ByteTrack \cite{zhang2022bytetrack}, all using the latest MegaDetector~(MD)~\cite{beery2019efficient} as the detector. Note that \samthree models trained on \dname were not exposed to the ``animal'' query during training; this prompt is used only at inference time. 

As shown in \Cref{tab:simple_baselines}, \samthree trained on \dname outperforms other methods by a substantial margin, surpassing the closest competing models on IDF1  and HOTA by +23.9 (row 4 vs. row 3) and +18.9 (row 4 vs. row 2), respectively.

\begin{table}[!t]
\small
\centering
\begin{tabular}{@{}clrrrrr@{}}
\toprule
 &  & \multirow{2}{*}{\textbf{\cgf}} & \multicolumn{3}{c}{\textbf{pHOTA}}  & \multirow{2}{*}{\textbf{\hspace{-4pt}TETA}}\\
\cmidrule(lr){4-6}
 & \hspace{-7pt}\textbf{Method} &  & \textbf{Total} & \textbf{Det} & \textbf{Ass} &\\
\midrule
1 & \hspace{-7pt}GLEE & -0.2 & 7.5 & 1.2 & 49.7 & 22.0 \\
2 & \hspace{-7pt}\makebox[75pt][l]{LLMDet + \samthree TR} & 2.6 & 41.3 & 21.4 & 80.0 & 30.4 \\
3 & \hspace{-7pt}\samthree & 14.0 & 48.5 & 28.4 & 83.4 & 39.6 \\
4 & \hspace{-7pt}\samthree (\dname) & 39.0 & 63.1 & 47.9 & 83.9 & 52.1 \\
5 & \hspace{-7pt}\makebox[75pt][l]{\samthree~FT~(\dname)} & \textbf{46.9} & \textbf{68.1} & \textbf{55.4} & \textbf{84.6} & \textbf{58.7} \\
\bottomrule
\end{tabular}
\caption{\textbf{Species Category-Specific Evaluation Results}. The inclusion of \dname during training (row 4) or fine-tuning (row 5) leads to a substantial improvement in performance over the baseline (row 3). The highest performance is shown in bold. See \Cref{sec:experiments}.}
\label{tab:results}
\end{table}

\begin{table}[!ht]
\footnotesize
\centering
{\fontsize{\mytablesize}{\mytablebaselineskip}\selectfont
\begin{tabular}{@{}x{4}y{90}x{17}x{20}x{20}x{20}@{}}
\toprule
 &  & \textbf{IDF1} & \multicolumn{3}{c}{\textbf{HOTA}} \\
 \cmidrule(lr){4-6}
 & \textbf{Method} && \textbf{Total} & \textbf{Det} & \textbf{Ass}\\
\midrule
1 & MD+ByteTrack   & 38.6 & 39.5 & 20.8 & 75.6 \\
2 & MD+OCSort      & 33.9 & 45.5 & 27.9 & 74.9 \\
3 & MD+BoostSort++ & 47.2 & 38.3 & 18.4 & 80.8 \\
4 & \samthree (\dname)        & \textbf{71.1} & \textbf{64.4} & \textbf{50.0} & \textbf{83.7} \\
\bottomrule
\end{tabular}
}
\caption{\textbf{Species Category-Agnostic Evaluation Results}. \samthree (\dname) surpasses all other models by large margins. \samthree was trained on the species-specific version of the training set. The highest performance is shown in bold. See \Cref{sec:experiments}.}
\vspace{-20pt}
\label{tab:simple_baselines}
\end{table}

\begin{absolutelynopagebreak}
\paragraph{\dname~Subset Evaluation}
Finally, we report the performance of the \samthree (SA-FARI) model using the species-specific prompting strategy on the splits described in \Cref{sec:dataset}. As shown in \Cref{tab:dataset_ablation}, samples containing smaller masklets are significantly harder to detect and track than those with larger masklets, as expected. The performance on large masks is +29.2 greater than small ones on the pHOTA metric. Masklets with occlusions and motion (in the ``challenging'' subset) are similarly difficult to detect, underperforming the overall set by -4.1 pHOTA-Det, and are also substantially harder to track (-9.3 pHOTA-Ass compared to the overall set). \end{absolutelynopagebreak}

The difficulty of samples with multiple animals is comparable to the average difficulty across the test set (\eg 67.8 vs. 68.1 pHOTA for multiple animals and the overall set, respectively). This is because, while detecting and tracking multiple animals is inherently more challenging owing to increased \textit{individual} ambiguity, these samples are positively correlated with large masks, which compensates for the increased tracking difficulty. Finally, nighttime videos present somewhat more challenging detection scenarios, which result in slightly lower scores, however, these results remain largely comparable to those from the rest of the test set (\eg 44.1 vs. 46.9 \cgf and 66.0 vs. 68.1 pHOTA for nighttime and overall set, respectively). Note that, as mentioned in \Cref{sec:dataset}, the night subset includes a majority of the videos in \dname/test.

\begin{table}[!t]
\footnotesize
\centering
{\fontsize{\mytablesize}{\mytablebaselineskip}\selectfont
\begin{tabular}{lccccc}
\toprule
 & \textbf{\cgf} & \multicolumn{3}{c}{\textbf{pHOTA}}  &  \textbf{TETA}\\
\cmidrule(lr){3-5}
\textbf{\dname/test split} &  & \textbf{Total} & \textbf{Det} & \textbf{Ass} &\\
\midrule
Large masks & 63.4 & 81.4 & 72.5 & 91.6 & 71.7 \\
Small masks & 25.3 & 52.2 & 38.5 & 71.3 & 46.9 \\
Multiple animals & 45.9 & 67.8 & 54.5 & 85.1 & 59.6 \\
Challenging & 36.8 & 61.7 & 51.3 & 75.3 & 59.6 \\
Night & 44.1 & 66.0 & 53.2 & 83.1 & 61.9 \\
All & 46.9 & 68.1 & 55.4 & 84.6 & 58.7 \\
\bottomrule
\end{tabular}
}
\caption{\textbf{Performance Across Test-Time Factors in \dname{}.} 
Samples with smaller masklets are significantly harder to detect and track than those with larger ones. 
Occluded or moving animals (``challenging'') are similatly hard to detect but notably harder to track. Videos with multiple animals exhibit moderate difficulty, offset by their tendency to contain larger masks. Nighttime samples yield overall performance comparable to the full test set.
}

\label{tab:dataset_ablation}
\end{table}
\vspace{-10pt}
\section{Conclusion}
\label{sec:conclusion}

We present \dname, a new large-scale dataset for MAT in the wild. By consolidating over 11,000 camera trap videos from over 700 sampling sites, \dname offers unprecedented scale and diversity, encompassing 99 species categories and sightings spanning a decade. We provide exhaustive, and -- for the first time -- manually verified spatio-temporal segmentation annotations at scale, resulting in over 16,000 reliable, high-quality masklets which are associated to individual animal identities. \dname thereby addresses a critical gap, as previous datasets are limited by narrow geographic and temporal scope, few species, and less reliable annotation methods.

Our results show that even SOTA models perform poorly in this domain without large-scale, richly annotated collections like \dname, underscoring the difficulties presented by real-world data sourced in the wild. Critically, incorporating our training split leads to substantial performance gains, with the fine-tuned \samthree model achieving a threefold improvement over its baseline. Analysis of our test splits highlights challenges, providing a roadmap for future data collection, annotations, and model development.

\dname can be extended, for instance by adding new modalities such as animal body pose, depth, and natural language descriptions. The integrated audio streams present a compelling opportunity for future multi-modal model development that leverage vocalizations to enhance detection, species classification and tracking robustness in the wild.

Future data integration should prioritize additional ecoregions to capture a wider array of species and mitigate geographic bias. By providing a rich, diverse, and rigorously annotated benchmark, we hope to accelerate progress toward the development of generalizable and robust tracking systems capable of enabling more effective and scalable biodiversity monitoring and protection across the globe.

\clearpage

{\footnotesize
\paragraph{Acknowledgements}
Osa Conservation acknowledges field support from staff members, volunteers and the Osa Camera Trap Network to collect and annotate their video data, and funding from Michael Simons and Sabrina Karklins, Biome Conservation, the Bobolink Foundation, the BAND Foundation, the Krystyna and Dan Houser Foundation, Troper Wojcicki Philanthropies, the KHR McNeely Family Fund, and the Mazar Family Charitable Foundation Trust. 

We thank Haitham Khedr and Ho Kei Cheng for their help with model evaluations, and Tengyu Ma for help with annotations.

We thank the Pan African Programme: ‘The Cultured Chimpanzee’ team and its collaborators for allowing the use of their data for this paper. We thank Amelie Pettrich, Antonio Buzharevski, Eva Martinez Garcia, Ivana Kirchmair, Sebastian Schütte, Linda Gerlach and Fabina Haas. We also thank management and support staff across all sites; specifically Yasmin Moebius, Geoffrey Muhanguzi, Martha Robbins, Henk Eshuis, Sergio Marrocoli and John Hart. Thanks to the team at https://www.chimpandsee.org particularly Briana Harder, Anja Landsmann, Laura K. Lynn, Zuzana Macháčková, Heidi Pfund, Kristeena Sigler and Jane Widness. The work that allowed for the collection of the dataset was funded by the Max Planck Society, Max Planck Society Innovation Fund, and Heinz L. Krekeler. In this respect we would like to thank: Ministre des Eaux et Forêts, Ministère de l'Enseignement supérieur et de la Recherche scientifique in Côte d’Ivoire; Institut Congolais pour la Conservation de la Nature, Ministère de la Recherche Scientifique in Democratic Republic of Congo; Forestry Development Authority in Liberia; Direction Des Eaux Et Forêts, Chasses Et Conservation Des Sols in Senegal; Makerere University Biological Field Station, Uganda National Council for Science and Technology, Uganda Wildlife Authority, National Forestry Authority in Uganda; National Institute for Forestry Development and Protected Area Management, Ministry of Agriculture and Forests, Ministry of Fisheries and Environment in Equatorial Guinea.
The authors would like to thank the Animal Biometrics group within the Machine Learning and Computer Vision (MaVi) research group at the University of Bristol for their valuable support. The work carried out by the group was partly supported by the UKRI Centre for Doctoral Training in Interactive Artificial Intelligence (CDT in Interactive AI) under grant EP/S022937/1.

We thank the Biodiversa+ project “Big\_Picture” (ref. Proyecto PCI2024-153504 convocatoria europea Biodiversa+, funded by MICIU/AEI) and Plan Nacional ref. PID2022-142919OB-100 for sharing their data for this paper.

We extend our sincere gratitude to Conservación Amazónica–ACCA for access to the Los Amigos Conservation Concession and for ongoing logistical support. We also thank the promotores (field rangers) of Los Amigos, whose dedication to maintaining the concession and whose assistance with camera trap deployment and monitoring made this work possible.
}
{
    \small
    \bibliographystyle{ieeenat_fullname}
    \bibliography{main}
}
\clearpage
\setcounter{figure}{0}
\setcounter{table}{0}
\setcounter{section}{0} \renewcommand{\thesection}{\Alph{section}}
\renewcommand{\thesubsection}{\thesection.\arabic{subsection}}
\renewcommand{\thefigure}{S\arabic{figure}}
\renewcommand{\thetable}{S\arabic{table}}
\renewcommand{\figurename}{Supplementary Figure}
\renewcommand{\tablename}{Supplementary Table}

\setcounter{page}{1}
\maketitlesupplementary
\section{Additional Dataset Statistics}
\label{sec:appendix_statistics}

\begin{figure*}[!b]
    \centering
    \includegraphics[width=\textwidth]{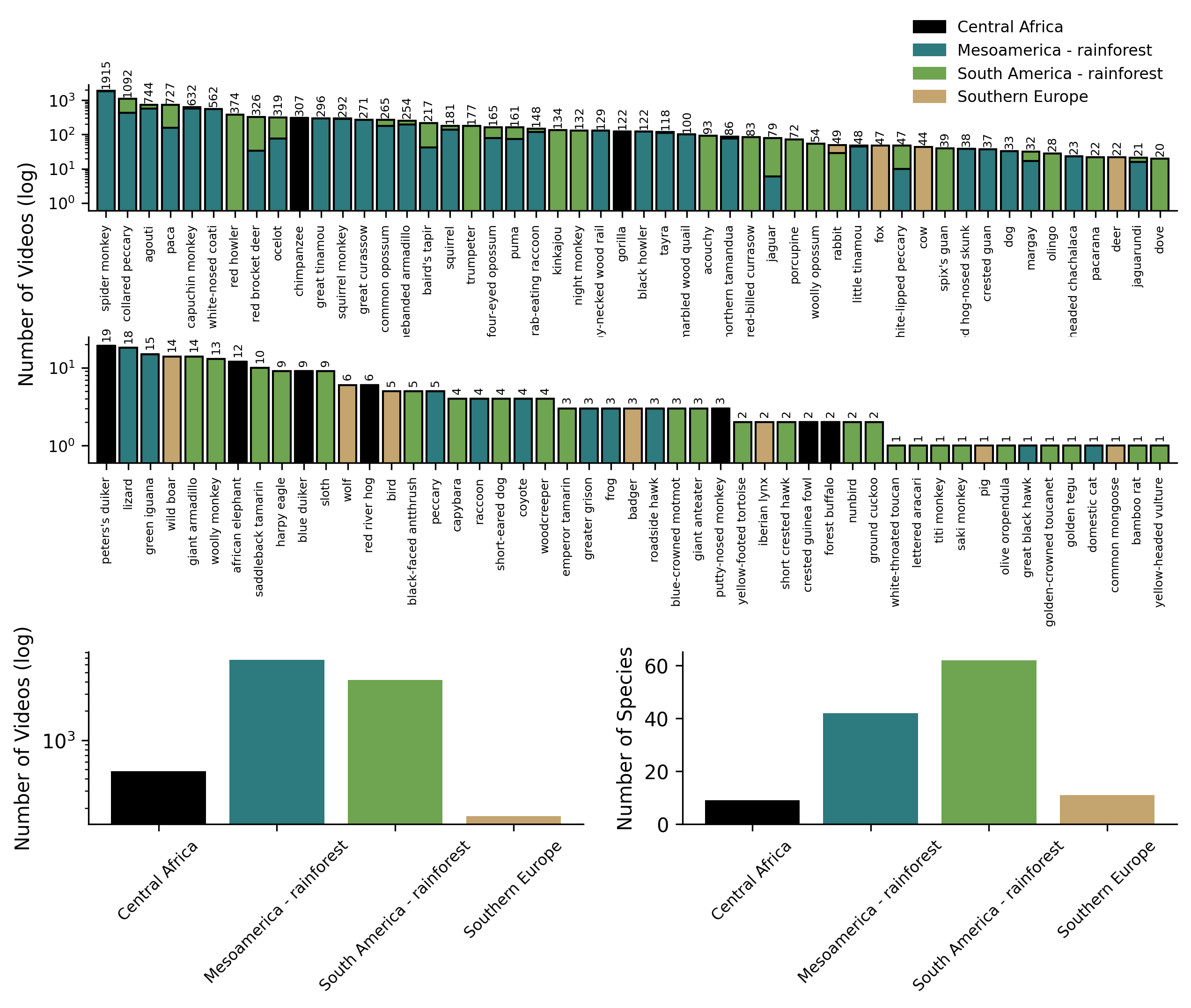}
    \caption{
        \textbf{Video and Species Category Distribution per \dname{} Ecoregion.} Video counts per species category and per continent.
    }
    \label{fig:datasets}
\end{figure*}

\begin{figure}[]
    \centering
    \includegraphics[width=1.\linewidth]{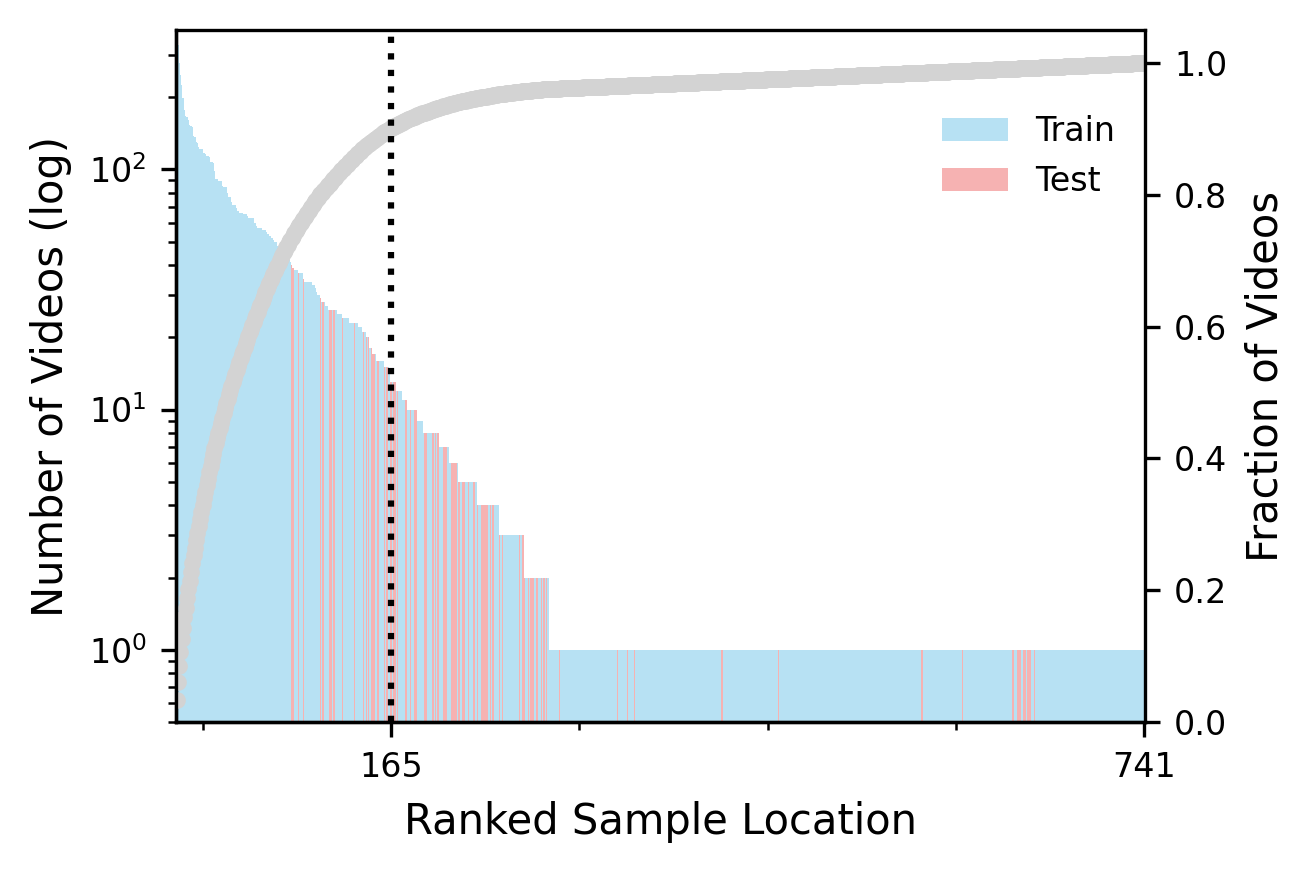}
    \caption{
        \textbf{Number of Videos per Camera Trap Location (Ranked).} Vertical line marks the minimum number of locations that account for 90\% of the videos. Note, train and test videos come from different locations.
    }
    \label{fig:ranked}
\end{figure}

\Cref{fig:datasets} shows the distribution of species annotations per continent, and \Cref{fig:ranked} shows the distribution of the number of videos per location.

\section{Metrics}
\label{sec:metrics}

In this section we provide an overview of the segmentation and tracking metrics we use in \Cref{sec:experiments}. For more details, we refer the reader to the \samthree \cite{sam3} paper.

\paragraph{IDF1} IDF1 measures the accuracy of maintaining object identities in multi-object tracking. It calculates the ratio of correctly matched detections (where both detection and identity are correct) to the average number of ground-truth and predicted detections, balancing precision and recall while penalizing identity switches.

\paragraph{HOTA} Higher Order Tracking Accuracy (HOTA) jointly evaluates detection and association in tracking. It decomposes performance into detection accuracy (DetA) and association accuracy (AssA), providing a balanced view of how well objects are detected and tracked over time \cite{luiten2020IJCV}.

\paragraph{pHOTA} Phrase-based HOTA (pHOTA) adapts HOTA for open-vocabulary tracking. Each video–noun phrase pair is treated as a unique sample, enabling class-agnostic evaluation. This is ideal for open-world settings, where tracked objects are specified by phrases rather than fixed categories.

\paragraph{TETA} Track Every Thing Accuracy (TETA) builds upon the HOTA metric, while extending it to better deal with multiple categories and incomplete annotations. It consists of three parts: a localization score, an association score, and a classification score \cite{trackeverything}.

\paragraph{\cgf} Classification-gated F1 (\cgf) is designed for open-vocabulary segmentation and tracking. It combines localization quality (positive micro F1, \pmf) and classification ability (image-level Matthews correlation coefficient, \ilmcc), rewarding models that are both accurate in localizing objects and calibrated in predicting their presence. This metric addresses the limitations of traditional AP in large label spaces.

\end{document}